\pdfoutput=1
\documentclass[11pt]{article}

\usepackage{acl}

\usepackage{times}
\usepackage{latexsym}
\usepackage{multirow}
\usepackage{tabularx} 
\usepackage{ragged2e}
\usepackage{graphicx} 
\usepackage{array}
\usepackage{booktabs} 
\usepackage{geometry}      % Optional: For adjusting page margins
\usepackage{caption}       % Optional: For caption customization
\usepackage{subcaption}
\usepackage{amsmath}
\usepackage{amssymb}
\usepackage{comment}
\usepackage{calc} 
\usepackage[T1]{fontenc}
\usepackage[utf8]{inputenc}
\usepackage{xspace}
\usepackage{microtype}

\usepackage{inconsolata}

\usepackage{arydshln}

\makeatletter
\def\adl@drawiv#1#2#3{%
        \hskip.5\tabcolsep
        \xleaders#3{#2.5\@tempdimb #1{1}#2.5\@tempdimb}%
                #2\z@ plus1fil minus1fil\relax
        \hskip.5\tabcolsep}
\newcommand{\cdashlinelr}[1]{%
  \noalign{\vskip\aboverulesep
           \global\let\@dashdrawstore\adl@draw
           \global\let\adl@draw\adl@drawiv}
  \cdashline{#1}
  \noalign{\global\let\adl@draw\@dashdrawstore
           \vskip\belowrulesep}}
\makeatother
\title{The Curious Case of Factual (Mis)Alignment \\ between LLMs' Short- and Long-Form Answers}

\author{%
  Saad Obaid ul Islam\textsuperscript{1} \quad
  Anne Lauscher\textsuperscript{2} \quad
  Goran Glavaš\textsuperscript{1} \\[1ex]
  \textsuperscript{1}WüNLP, CAIDAS, University of Würzburg \\ \texttt{\{saad.obaid-ul-islam,goran.glavas\}@uni-wuerzburg.de} \\
  \textsuperscript{2}Data Science Group, University of Hamburg \\ \texttt{anne.lauscher@uni-hamburg.de}
}
\newcommand{\rparagraph}[1]{\vspace{1.2mm}\noindent\textbf{#1}}

\newcommand{\sparagraph}[1]{\vspace{0.0mm}\noindent\textbf{#1}}
\newcommand{\acrnm}{\texttt{SLAQ}\xspace}
\begin{document}
\maketitle
\begin{abstract}

Large language models (LLMs) can correctly answer \textit{"When was Einstein born?"} yet fail to provide the same date when the same question is embedded in a long-form query requesting multiple facts about Einstein's life, revealing a fundamental inconsistency in how models access facts across different complexities of knowledge seeking tasks. While models display impressive accuracy on factual question-answering benchmarks, the reliability gap between simple and complex long-form queries remains poorly understood, eroding their trustworthiness. 
In this work, we introduce Short-Long Form Alignment for Factual Question Answering (\acrnm), a controlled evaluation framework that compares LLMs' answers to the same factual questions asked (a) in isolation (\textit{short}) vs. (b) integrated into complex queries (\textit{long}). Looking at 16 LLMs across 600 queries, we find a systematic misalignment of answers to the corresponding \textit{short} and \textit{long} queries. 
We further uncover momentum effects where consecutive correct or incorrect answers create self-reinforcing patterns. Through mechanistic analysis, we find that aligned facts activate overlapping model internals, and that metrics based on mechanistic similarity can predict \textit{short}-\textit{long} answer alignment with up to 80\% accuracy. Our work establishes \textit{factual consistency over query complexity} as an important aspect of LLMs' trustworthiness and challenges current evaluation practices, which implicitly assume that good performance for simple factual queries implies reliability in more complex knowledge-seeking tasks as well.   

\end{abstract}
\section{Introduction}\label{sec:intro}

\begin{figure*}[h]
    \centering
    \includegraphics[width=\textwidth]{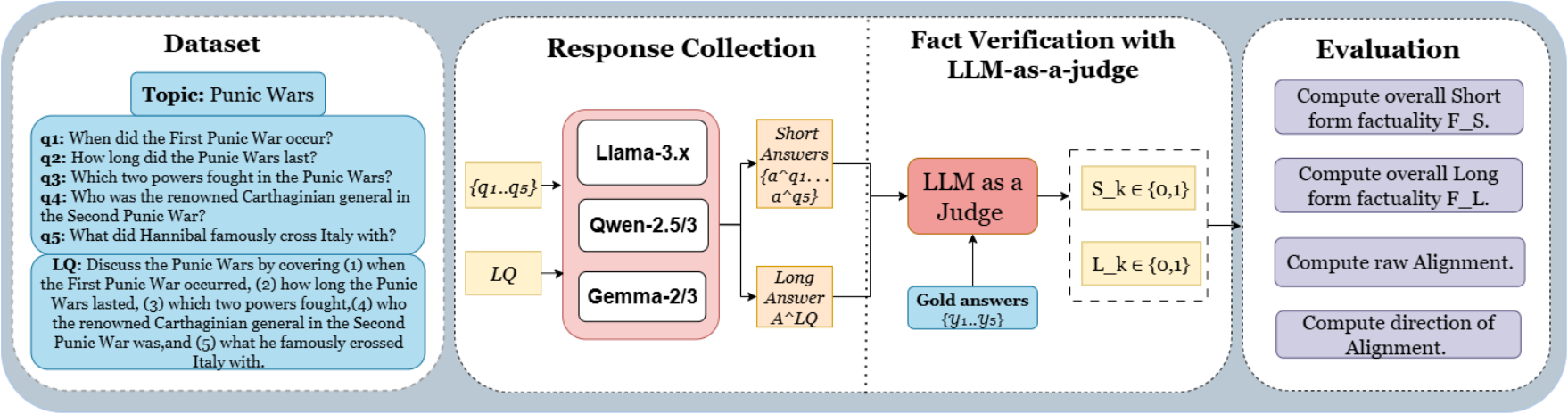}
    \caption{Illustration of our \textbf{Short-Long Form Alignment for Factual Question Answering} (\acrnm) framework. An instance in our \acrnm benchmark is a \textit{complex} knowledge-seeking query, i.e., a \textbf{long} query, which consists of five \textit{simple} factual sub-queries, i.e., \textbf{short} queries, each with an unambiguous correct answer.  %  consisting of five dataset provides five short questions, one composed long question, and gold answer sets. 
    LLMs independently generate the answers to (1) the \textbf{long} query (i.e., all five short queries combined) and (2) each of the five short queries in isolation. 
    %Models answer the shorts independently and the long question once. 
    We use a state-of-the-art commercial LLM to judge the correctness of the generated answers to both the \textit{long} query and \textit{short} queries against the set of reference answers; we use these judgments 
    %and returns binary correctness labels for short and long. 
    to compute models' short- and long-form accuracy ($F_{S}$, $F_{L}$) as well as the \textit{short-long alignment} scores.
    %($Align$) that checks whether the two formats agree on each fact, and a Signed Alignment score ($Align_{\pm}$) that indicates the direction of alignment.
    }
    \label{fig:pipe-line}
    \vspace{-0.5em}
\end{figure*}

Large Language Models (LLMs) \citep{team2023gemini, achiam2023gpt, grattafiori2024llama, yang2025qwen3} are rapidly being adopted across diverse applications, including education \cite{kasneci2023chatgpt}, healthcare \cite{qiu2023large}, software engineering \cite{fan2023large}, and general knowledge search \cite{xu2023chatgpt}. Their utility and trustworthiness are, however, compromised by their tendency to hallucinate \cite{Wang2023SurveyOF} and generate fictitious responses \cite{huang2025survey}.

While earlier research on LLM evaluation extensively examined factual accuracy in closed-domain question answering (QA) for both short-form \cite{rajpurkar2016squad, joshi2017triviaqa, yang2018hotpotqa} and long-form responses \cite{fan2019eli5, dasigi2021dataset}, these evaluation benchmarks have become somewhat outdated, since LLMs are now primarily deployed as chat-based information gathering assistants across a wide variety real-world applications \cite{xu2023chatgpt}, i.e., they are primarily used as (chat-based) open-domain question-answering tools.
Accordingly, factuality-oriented evaluations have shifted toward open-domain QA, considering both short-form \cite{lin2022truthfulqa, wei2024long} and long-form responses \cite{min2023factscore, wei2024long, Islam2025HowMD}. 
%, as this better reflects how LLMs are actually used. 
Existing benchmarks, however, evaluate short-form and long-form factuality in isolation, and thus fail to assess \textit{factual consistency} of models' responses over query complexity: \textit{Will an LLM yield the same answer to the same factual question for queries of varying complexity?}

In this work, we address this gap by introducing Short-Long Form Alignment for Factual Question Answering (\acrnm), a novel evaluation framework that tests whether models maintain answer consistency---with respect to fact-seeking questions---across queries of different complexity. 
\acrnm presents an LLM with the same fact-seeking questions, formulated (independently) in two distinct query formats: (1) \textit{long} queries combine five topically related factual questions, whereas (2) \textit{short} queries formulate those same questions independently and ask them in isolation. 
%%
%presents identical factual content through two formats: as independent short-form questions and as components within an integrated long-form question. 
With this controlled design, we isolate the impact of query/response complexity on factual answer accuracy. 
By comparing the factual correctness of models' answers to long vs. short queries, %performance across these matched conditions, 
we can disentangle knowledge gaps (incorrect answers for both query formats) from answer retrieval failures (e.g., correct answer for the short query but incorrect for the long). Figure \ref{fig:pipe-line} illustrates \acrnm.

%%%
%%% GG stopped here
Studying 16 LLMs through the lens of \acrnm, we find that, while models exhibit substantial \textit{short-long alignment} w.r.t. factual answer correctness, most of this alignment stems from incorrect answers, i.e., LLMs produce \textit{incorrect} answers for the same factual question in both long and short-form queries (but it is not necessarily the same answer).  
%answers to the fact-seeking questions.
%primarily when both responses are 
%incorrect. 
We observe that models consistently demonstrate higher factual accuracy in responses to short queries than in long-query responses: the majority of misalignment cases thus stem from a (1) correct answer to the short query and an (2) incorrect answer to the corresponding question included in the \textit{long} query.  
%
%it is predominantly driven by the lower factuality of long-form responses compared to their short-form counterparts. , we observe that models consistently demonstrate higher accuracy in short-form responses than in long-form responses.
%
Beyond evaluating factual question answering accuracy for both query formats, we identify momentum effects, where consecutive correct answers increase the likelihood of subsequent accuracy, while errors tend to cascade and compound.

To understand the mechanistic basis of the observed factual misalignment, we next analyze the model internals--- attention and MLP activation patterns---and identify minimal sets of model components responsible for answer generation for short and long-form queries, respectively. Using zero-ablation \cite{olsson2022context} activation patching \cite{meng2022locating}, we find that aligned answers activate significantly more similar computational pathways and exhibit stronger correlations in component importance rankings. 
Moreover, we show that these circuit-level differences have predictive power: employing six pathway similarity metrics, we can predict with 80\% accuracy (ROC-AUC: 0.85) whether the answers to the same factual question will align between the two query formats, short and long; here we identify attention head rank correlation as the most predictive feature.

\sparagraph{Contributions.} In sum, the contributions of this work are threefold: (1) We establish \textit{factual consistency over query complexity} as an important aspect of LLM reliability and introduce \acrnm, a novel dataset for benchmarking such consistency; (2) We document systematic factual misalignment (i.e., inconsistency) patterns in LLMs, and relate factual correctness of the responses to position effects and momentum dynamics; (3) We provide mechanistic evidence that this factual misalignment stems from divergent internal processing, demonstrating that circuit overlap metrics can predict alignment outcomes.
This work represents the first systematic investigation of factual consistency over query complexity in open-domain QA. Our findings challenge a fundamental (implicit) assumption of modern LLM evaluation: that factual knowledge that LLMs exhibit for simple queries with straightforward factual questions propagates reliably to complex scenarios, where the same factual questions are part of more complex knowledge-seeking queries\footnote{Github: https://github.com/WorldHellow/SLAQ}.

\section{Background and Related Work}

We provide a brief overview of background and related work on (1) hallucinations and factuality in open-domain QA, and (2) mechanistic interpretability and its application to understanding factuality.

\rparagraph{Hallucinations and Factuality.} Evaluating factual accuracy in LLMs has evolved from simple to complex formats. Early benchmarks like TriviaQA \citep{joshi2017triviaqa} and Natural Questions \cite{kwiatkowski-etal-2019-natural} evaluated LLMs on closed-domain QA. But these benchmarks are now saturated, and evaluation of LLMs has moved from closed-domain to open-domain QA, with TruthfulQA \cite{lin2022truthfulqa} and SimpleQA \cite{wei2024measuring} as two popular benchmarks that evaluate LLMs for single factoid answers.

With respect to factual accuracy in long-form LLM responses, FactScore \cite{min2023factscore} evaluates LLMs on Wikipedia biographies. More recently, LongFact \cite{wei2024long} and UNCLE \cite{yang2025hallucinate} were proposed to evaluate long-form factual accuracy across diverse domains. UNCLE is a concurrent effort to ours and, similarly to our work, pairs short and long queries/prompts: however, it analyses the short- and long-form in isolation and studies uncertainty expression rather than factual consistency of LLMs' responses between the two query formats. 

Several systematic phenomena have been observed regarding hallucinations in LLMs. The ``snowballing" effect \cite{zhang2024how} describes how models justify a wrong claim by generating additional false assertions. The ``lost in the middl'' phenomena \cite{liu-etal-2024-lost} shows input-position sensitivity in closed-domain QA: accuracy peaks when evidence appears at the beginning or end of a long context and degrades when relevant information lies in the middle. Complementing input-position effects, \citet{yang2025hallucinate} find that hallucinations in long-document summarization occur more often near the end of generated outputs (``hallucinate at the end''), indicating degradation dependent on the output position. 

\rparagraph{Mechanistic Interpretability} (MI) aims to reverse-engineer how neural networks result in specific behaviors by identifying causal computational structures \cite{elhage2021mathematical,zhang2024how}. The foundation of MI is localization: determining which model components (attention heads, MLP layers, neurons) are responsible for particular outputs. The primary technique for measuring component importance is activation patching \cite{meng2022locating}, which quantifies causal influence through intervention. The process involves: (1) computing baseline output logits $\ell_{\text{base}}$ for the correct token, (2) ablating each component individually to obtain $\ell_{\text{ablated}}$, (3) measuring the importance as normalized logit difference: $|\ell_{\text{base}} - \ell_{\text{ablated}}|/|\ell_{\text{base}}|$, where larger values indicate greater causal importance. Components are then assembled into minimal circuits through greedy search \cite{conmy2023towards,hanna2024have}—iteratively adding components in importance order until the subset reproduces the original behavior within a faithfulness threshold \cite{wang2023interpretability}. Two ablation strategies exist: zero-ablation \cite{olsson2022context} sets component outputs to zero, while counterfactual patching replaces them with activations from different inputs \cite{meng2022locating}. In this work, for computational efficiency, we resort to zero-ablation when identifying component sets.

A significant amount of work focused on identifying parameters in which models store factual information. ROME \cite{meng2022locating} localizes factual associations to mid-layer MLPs, for which \citet{geva-etal-2023-dissecting} further show that they function as key-value memories. \citet{yao2024knowledge} trace factual retrieval circuits, revealing collaborative knowledge encoding across attention heads and MLPs. Limited work exists on comparing circuits between tasks: \citet{mondorf2024circuit} find high node overlap for compositionally similar tasks, while \citet{hanna2025formal} report minimal overlap between formal and functional linguistic circuits. These studies share a critical limitation: they analyze single-token outputs, ignoring the complexity of realistic free-form multi-token answers. 

\section{Factual Consistency over Query Complexity}\label{sec:task}

Our goal is to capture the extent to which LLMs provide consistent (i.e., factually equivalent) answers to the very same fact-seeking questions, integrated into queries of different complexity. To this end, we introduce a novel task of factual answer consistency over query complexity, for which we create an evaluation benchmark.      
%first describe the task on which LLMs will be evaluated and metrics utilized to evaluate LLM responses. And then we describe how we construct the datasets and provide dataset statistics.

\subsection{Task Definition and Metrics} \acrnm tests whether LLMs provide consistent answers to factual questions across different query/response complexities. Because of this, we organize the benchmark around \textit{topics}: a topic $t$ is a set of $N$ topically related facts $\{f_1, f_2, \dots, f_N\}$, for each of which \acrnm contains a \textit{short}-form question (\texttt{SQ}) that elicits the respective fact, $\{q_1, q_2, \dots, q_N\}$. Each topic, as a set of $N$ facts, is additionally converted into a \textit{long} information-seeking (\texttt{LQ}) query: an example of a topic $t$ with $N=5$ factual questions is given in Figure~\ref{fig:pipe-line}. The LLMs then independently respond to the \texttt{LQ}, as well as to each of the $N$ \texttt{SQs}. 
%
%(e.g., ``Who started the Punic wars'')    
%Each topic in our dataset contains 5 independent facts about the topic, 5 short questions (one per fact), and 1 long question requesting all 5 facts together. Models answer each short question individually and the long question comprehensively. 
For each factual question $q_k$ ($k \in \{1, 2, \dots, N\}$), $S_k \in \{0, 1\}$ denotes the factual correctness of an LLM's answer to the \texttt{SQ} of that fact (1 = correct, 0 = incorrect) whereas the $L_k \in \{0, 1\}$ indicates the correctness for the same fact in the LLM's answer to the \texttt{LQ}. 

\rparagraph{Alignment Definition.} 
%%%
We declare that an LLM produces a \textit{factually consistent} response for a fact $f_k$ if the \texttt{SQ} and \texttt{LQ} responses for that fact have the same factual correctness label: %A fact is aligned when it receives the same correctness label in both responses: 
\begin{equation}
    \footnotesize
    \text{aligned}(k) = \mathbb{I}\{S_k = L_k\}
\end{equation}
\noindent where $\mathbb{I}\{\cdot\}$ is the indicator function. Crucially, alignment measures the consistency in factual correctness of the answers, and not whether the answers themselves are semantically equivalent. When both responses are incorrect ($S_k = L_k = 0$), they are aligned w.r.t. factual correctness because they are both incorrect. 
%even if stating different wrong answers. 
E.g., for question $q_1$ from Figure~\ref{fig:pipe-line}, answers ``264 to 243 BCE'' (short) and ``264 to 241 AD'' (long) are factuality-wise aligned (both incorrect) because the correct answer is ``264 to 241 BCE''
%---both are wrong, though differently so. 
We choose this alignment definition as we aim to discern between (1) a knowledge gap (i.e., both answers incorrect; irrelevant if they are ``the same incorrect'') and (2) failure to consistently retrieve the knowledge that is stored in the model (i.e., one answer correct, the other incorrect).

\rparagraph{Evaluation Metrics.} 
Following prior work \citep{min2023factscore, wei2024measuring, wei2024long}, we an LLM to judge factual correctness by comparing responses against gold answers. We characterize model behavior with following metrics: Short and Long form factual \textbf{Accuracy} ($F_S$ and $F_L $), $Align$ment score, and \textbf{Signed Alignment} score ($Align_{\pm}$).
\begin{equation}
    \scriptsize
    F_S = \frac{1}{N}\sum_{k=1}^{N} S_k 
    \qquad
    F_L = \frac{1}{N}\sum_{k=1}^{N} L_k 
    \label{eq:FS_FL}
\end{equation}
\begin{equation}
    \scriptsize
    \text{Align} = \frac{1}{N}\sum_{k=1}^{N} \mathbb{I}\{S_k = L_k\}
    \label{eq:align}
\end{equation}
\begin{equation}
    \scriptsize
    \text{Align}_{\pm}=\tfrac{1}{N}\sum_{k=1}^{N}A_k,\quad
    A_k=\left\{\begin{array}{@{}r@{\,}l@{}}
    \ \ 1 & \text{if } S_k=L_k=1,\\[-1pt]
    -1 & \text{if } S_k=L_k=0,\\[-1pt]
    \ \ 0 & \text{if } S_k\neq L_k.
    \end{array}\right.
    \label{eq:signed-align}
\end{equation}

\noindent $F_S$ and $F_L $ establish baseline performance and quantify the accuracy gap between formats. $Align$ measures factual consistency regardless of correctness. Raw alignment score conflate reliable knowledge (both correct) and systematic failure (both incorrect), which is why we introduce $\text{Align}_{\pm}$, which additionally distinguishes the two correctness cases: +1 (aligned, correct) vs. $-1$ (aligned, incorrect), with $0$ denoting factual misalignment.

\begin{figure*}[t]
  \centering
  \includegraphics[width=\textwidth]{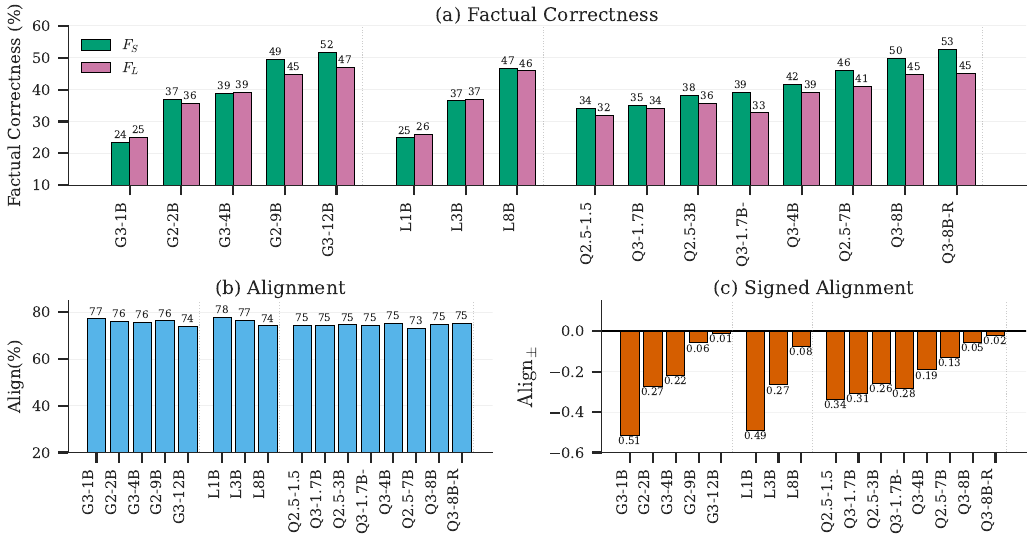}
  \caption{Short–long factual alignment results across model families. 
(a) \textbf{Factual Correctness}: per-model short-form accuracy \(F_S\) (green) and long-form accuracy \(F_L\) (purple). 
(b) \textbf{Alignment}: \(\mathrm{Align}\) = percentage of facts with the same correctness label in short vs.\ long responses. 
(c) \textbf{Signed Alignment}: average over topics; for a single topic, the score is the average of $\mathrm{Align}_{\pm}$ of its five facts.
Models key: G = Gemma, L = Llama, Q = Qwen (e.g., Q3–8B-R = Qwen-3, 8B parameters, R - reasoning).}
  \label{fig:slfa-metrics}
  \vspace{-0.5em}
\end{figure*}

\subsection{Dataset}\label{subsec:dataset} 
We construct \acrnm datasets from Wikipedia, leveraging its factual reliability and broad coverage. We sample from 15 diverse English Wikipedia categories, selecting articles exceeding 1,000 words to ensure sufficient factual density. To test models across the knowledge popularity spectrum, we balance between popular and obscure topics, selecting 300 most-viewed and 300 least-viewed pages in the past five years. This way, we take into account evidence \cite{zhang2025law} that fact frequency drives LLMs' hallucination. 

Following the success of LLM as synthetic data generators \cite{long-etal-2024-llms} in open-domain QA \cite{wei2024long, Islam2025HowMD, yang2025uncle} we employ a state-of-the-art commercial LLM, OpenAI \texttt{o3-mini-high}, to generate factual questions to which an answer exists in the article content. The model receives the full Wikipedia text and produces $N=5$ \texttt{SQ}s targeting distinct facts, plus one \texttt{LQ} that naturally elicits all five facts\footnote{We provide the prompts for generating \texttt{SQ}s, \texttt{LQ}s, and dataset samples in the \S\ref{subsec:appendix-dataset}}. 

We then manually verified all generated \texttt{SQ}s and \texttt{LQ}s, ensuring their open-domain formulation (i.e., that they are answerable without the source article) as well as factual grounding (the correct answer indeed exists in Wikipedia). We manually adjusted the queries that did not meet both criteria.\footnote{E.g., we identified 53 \texttt{SQ}s and 24 \texttt{LQ}s that violated open-domain criteria and were judged as unanswerable without the respective Wikipedia article}. 
%All the facts were grounded in the wikipedia article. 
Overall, we found \texttt{o3-mini-high} to be a reliable synthetic data generator for this purpose: it introduced errors in only 77 out of 3,600 (2.14\%) query-answer pairs.  

The \acrnm datasets covers 600 unique topics (with 600 corresponding \texttt{LQ}s) with 3000 \texttt{SQ}s ($N=5$), across 15 wikipedia categories (on average, 40 topics per category). We further profile the \texttt{SQ}s for fact-type, finding 1,071 entity-based facts and 1,929 non-entity facts (definitions, properties, equations, concepts)\footnote{For this labeling, we resort to the OntoNotes taxonomy.}
%(implemented in spaCy.}. 
The final \acrnm evaluation dataset is thus both category-diverse and popularity-balanced, and has a fair balance between factual questions with entity vs. non-entity target answers. 
\begin{figure*}[t]
  \centering
  \includegraphics[width=\textwidth]{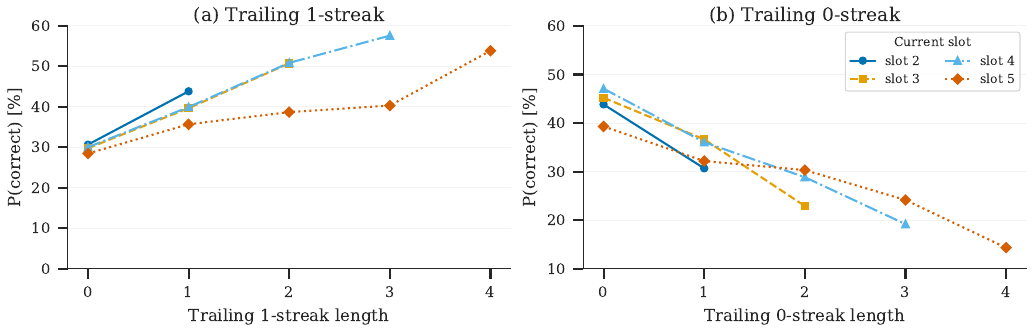}
  \caption{Long-form QA dynamics by sub-fact position. 
    (a) \textbf{Trailing 1-streak}: \(P(\mathrm{correct})\) for the current slot (2–5), conditioned on the length of the immediately preceding run of correct slots. 
    (b) \textbf{Trailing 0-streak}: \(P(\mathrm{correct})\) for the current slot (2–5), conditioned on the length of the immediately preceding run of incorrect slots.
    Slots (1-5) are sub-fact positions in the long-form query. 
    }
  \label{fig:long-form-analysis-panel}
  \vspace{-0.5em}
\end{figure*}

\section{Benchmarking and Evaluations}\label{sec:benchmarking}

\subsection{Experimental Setup}
We evaluate models from five families, spanning 1B to 12B parameters: Qwen-2.5 \citep{Yang2024Qwen25TR}, Qwen-3 and Qwen-3-Reasoning \citep{yang2025qwen3}, Llama-3 \citep{grattafiori2024llama}, Gemma-2 \citep{team2024gemma}, and Gemma-3 \citep{team2025gemma}. All models use greedy decoding via the Hugging Face API. We constrain short-form responses to single sentences and instruct models to provide only requested information for long-form queries (see Table~\ref{tab:instruct-short-long} in the Appendix for prompts). To control for positional effects, we randomize the order of the five sub-questions within each long-form query across 5 permutations and report averaged results.

We employ Gemini-2.5-Flash \citep{team2023gemini} as our LLM judge, instructing it to judge answer correctness based on semantic equivalence with the gold answer rather than string matching. The LLM judge agrees with the human annotator for 92.0\% and 94.8\% of \texttt{SQ} and \texttt{LQ} responses, respectively (see Appendix Tables~\ref{tab:prompts-short-form-eval} and~\ref{tab:prompts-long-form-eval} for prompts). We then compute our evaluation metrics (Eq.~\ref{eq:FS_FL}--\ref{eq:signed-align}) from the judge's binary correctness labels. 
%we compute our evaluation metrics , alignment (Eq.~\ref{eq:align}), and signed alignment (EQ.~) metrics.

\subsection{Results and Analysis}

Figure~\ref{fig:slfa-metrics} reveals three key patterns of factual (in)consistency across the language models. Panel (a) shows that most models achieve modest factual accuracy of 30-50\% on both \texttt{SQ} ($F_S$) and \texttt{LQ} ($F_L$) responses, with almost all models displaying higher accuracy for short-form queries. Larger models display only modestly better performance, suggesting that scale alone cannot dramatically improve factual recall.

Panel (b) shows remarkable consistency in raw alignment across all models, with virtually every model achieving 73-78\% alignment regardless of size and architecture. Such uniformity indicates that this level of factual consistency over query complexity is an intrinsic property of modern LLMs, rather than something that can improve with scale.

Panel (c) reveals the most critical finding: all models show negative signed alignment (-0.01 to -0.51), meaning they consistently provide wrong answers in responses to both query formats more often than correct answers for both cases. This indicates that the high raw alignment primarily reflects systematic failures rather than systematic successes: models have developed stable internal strategies for factual processing, but these strategies systematically fail to retrieve correct information. The combination of high raw alignment with negative signed alignment reveals that while models are internally consistent in factual behavior, most of it stems from generating incorrect answers across query complexities. %  consistency is predominantly in the wrong direction..

Panel (a) shows that $F_S$ is consistently larger than $F_L$.\footnote{The only exceptions are the two smallest models in our evaluation, Gemma-3 1B and Llama-3 1B, which yield very low accuracy for both \texttt{SQ}s and \texttt{LQ}s.} To better understand why $F_L$ is lower, we next analyze \texttt{LQ} responses in more detail.

\rparagraph{Momentum within a response.} According to Figure~\ref{fig:long-form-analysis-panel}b and~\ref{fig:long-form-analysis-panel}c, following consecutive correct answers (positive momentum), accuracy increases from a ~30\% baseline to 57\% after four correct facts: each additional success adds roughly 7\% to subsequent accuracy. Conversely, consecutive errors (negative momentum) reduce accuracy from ~45\% to 24\% after three mistakes. This not only confirms \citet{zhang2024how}'s finding of ``snowballing'' for long-form QA but extends it by quantifying the error propagation and success reinforcement effects. 
These momentum dynamics explain why \texttt{LQ} responses systematically underperform short-form ones even when eliciting the very same factual knowledge.

\section{Mechanistic Analysis}\label{sec:mechanistic-analysis}

Our behavioral analysis revealed that language models exhibit systematic inconsistencies when answering the same facts across short and long response formats. This raises a fundamental question: do these behavioral differences reflect distinct internal computational mechanisms? Understanding the mechanistic basis of factual alignment could inform interventions to improve consistency across response formats.
We hypothesize that factual alignment corresponds to mechanistic similarity. Formally, let $\text{sim}(k)$ denote the mechanistic similarity between short and long responses for fact $k$. Our hypothesis predicts:
\begin{equation}
    \footnotesize
    \mathbb{E}[\text{sim}(k) \mid aligned] > \mathbb{E}[\text{sim}(k) \mid misaligned]
\end{equation}
\noindent with alignment defined as in\,\S\ref{sec:task}. We focus exclusively on facts that are answered correctly in both formats (aligned) versus facts where only one format is correct (misaligned). Put simply: facts answered correctly in both short and long formats should exhibit greater internal mechanism overlap than facts answered correctly in only one format.

\begin{figure*}[t]
  \centering
  \includegraphics[width=\textwidth]{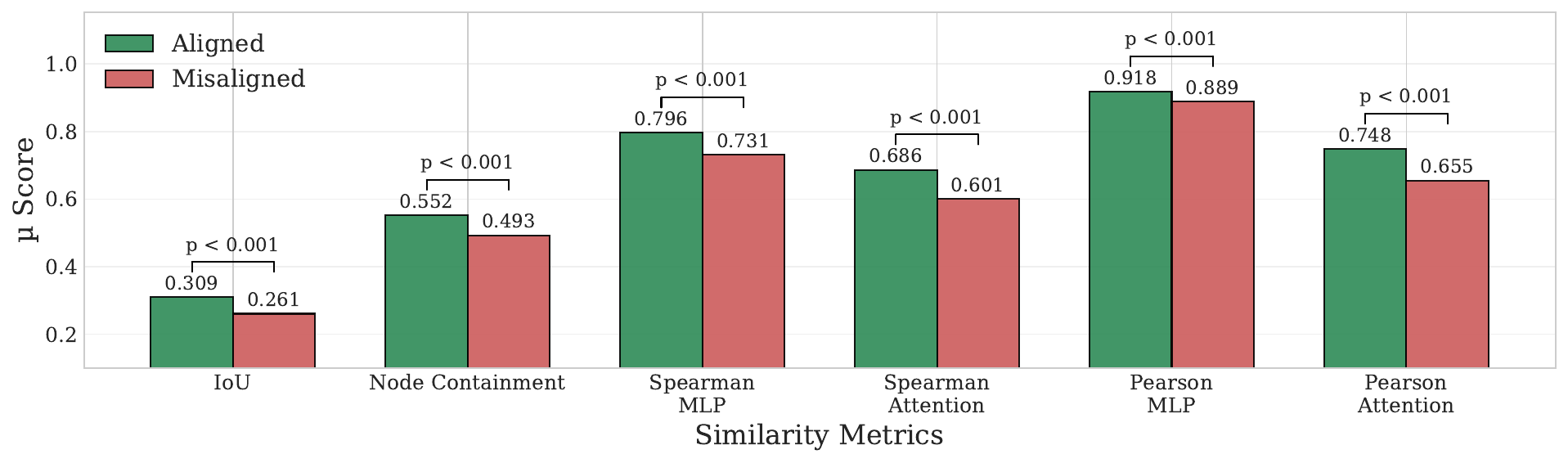}
  \caption{Mechanistic similarity comparison between aligned and misaligned facts across six metrics averaged across six LLMs. Aligned facts (green) show significantly (p < 0.05) higher mechanistic similarity than misaligned facts (red) for all measures. Scores are averaged across six LLMs.}
  \label{fig:circuit-metrics}
  \vspace{-0.5em}
\end{figure*}

\subsection{Preliminaries: Component Importance via Zero-Ablation}
We identify critical components by systematically setting their outputs to zero and measuring how much that hurts the model’s preference for the gold token.  For each component $c$ and answer token $t$, we measure importance using:
\begin{equation}
\small
\text{importance}(c, t) = \frac{\text{logit}_t^{\text{base}} - \text{logit}_t^{\text{ablated}}}{\text{logit}_t^{\text{base}}}
\end{equation}
\noindent This quantifies the change in logit magnitude when a component $c$ is removed, normalized by the baseline logit. For \textit{each} answer token, components are ranked by importance. We greedily select components (highest importance first) until we recover at least 90\% of the baseline logit, yielding the minimal set $C_t$ needed for generating token $t$. 

\paragraph{Similarity Metrics.}
We compare component sets of responses to \texttt{SQ}s and \texttt{LQ}s with two metrics:
\begin{equation}
    \small
    \text{Containment} = \frac{|C_{\text{short}} \cap C_{\text{long}}|}{\min(|C_{\text{short}}|, |C_{\text{long}}|)}\\
\end{equation}
\begin{equation}
    \small
    \text{Intersection-over-Union} = \frac{|C_{\text{short}} \cap C_{\text{long}}|}{|C_{\text{short}} \cup C_{\text{long}}|}
\end{equation}
\noindent with $C_{\text{short}}$ and $C_{\text{long}}$ being the component sets for short and long responses, respectively. Containment measures core component sharing relative to the smaller circuit, while IoU quantifies the overlap symmetrically. We additionally measure \textbf{Pearson Correlation} and \textbf{Spearman Correlation} between the two sets of component importance scores: the former captures the extent to which importance scores match/deviate across components, whereas the latter quantifies the extent to which the two rankings of components (by decreasing importance score) match.
%over the components rank correlation of component importance scores, which captures similar attention/MLP hierarchies; and \textbf{Pearson Correlation:} a linear correlation of importance scores, which captures magnitude relationships.

\paragraph{Multi-Token Alignment via Earth Mover's Distance (EMD)}

Previous studies on LLMs' factual revall \cite{meng2022locating, geva-etal-2023-dissecting, yao2024knowledge} focused on single-word answers. However, real answers to factual questions span multiple tokens, with variable lengths. Because we extract component sets per token, we need to aggregate token-level component-based similarity scores into fact-level (i.e., answer-level) scores, taking into account different token boundaries (e.g., \textit{``Paris is capital of France''} vs \textit{``the capital city Paris is located in France''}).
We formulate component comparison across multi-token answers as an optimal transport problem: we compute pairwise similarities between all \texttt{SQ} answer tokens $\{s_i\}$ and corresponding \texttt{LQ} answer tokens $\{l_j\}$, creating a bipartite similarity matrix $M_{ij}$. EMD then finds the transport plan matrix $\pi^*$ (i.e., coefficients $\pi_{ij}$) that maximizes the following total similarity:

\begin{equation}
\pi^* = \arg\max_\pi \sum_{i,j} \pi_{ij} \cdot M_{ij}
\end{equation}

\noindent The final fact-level similarity score is then the weighted average of pairwise similarities with optimal transport weights: $\text{sim}(k) = \sum_{i,j} \pi^*_{ij} \cdot M_{ij}$. Unlike exact matching, EMD finds the best possible alignment of tokens between multi-token answers, reflecting semantic equivalence between tokens regardless of their order in the answer.

\begin{figure*}[t]
  \centering
  \includegraphics[width=\textwidth]{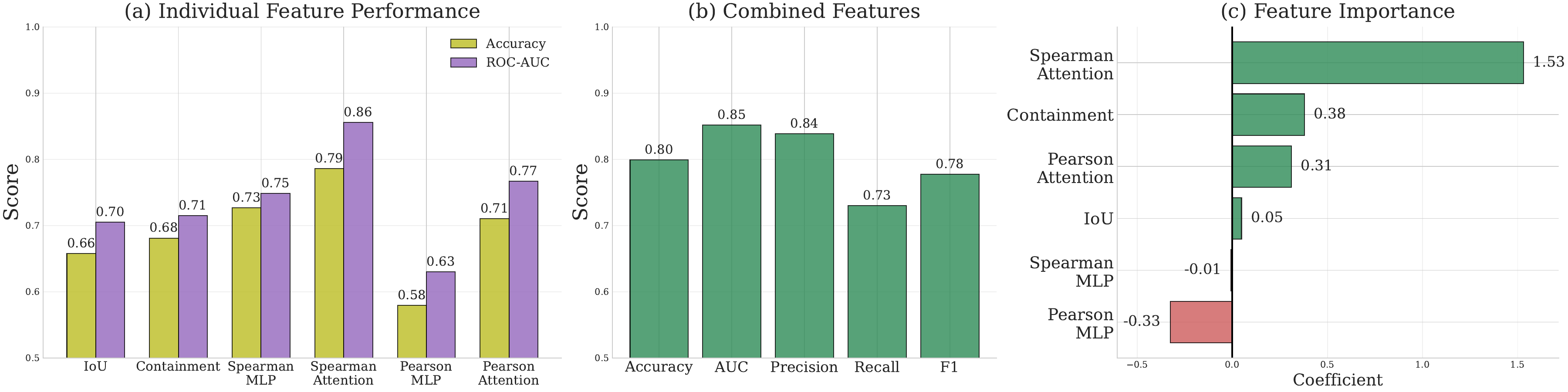}
  \caption{Predictive modeling performance using circuit similarity metrics. (a) Individual feature performance shows Spearman Attention as the strongest single predictor of factual alignment (ROC-AUC = 0.86). (b) Combined features achieve robust performance across all evaluation metrics (ROC-AUC = 0.85, Accuracy = 0.80). (c) Feature importance reveals Spearman Attention as the dominant predictor (coefficient = 1.53).}
  \label{fig:predictive-modeling}
\end{figure*}

\subsection{Mechanistic Comparison}

\rparagraph{Experimental Setup} We analyze four models spanning different scales: Qwen-2.5 (1B, 3B) and Qwen-3 (1.7B, 4B).  For each model, we select 60 short-long fact pairs that are divided into 30 correctly aligned pairs and 30 misaligned pairs. This yields us 240 short-long fact pairs.  For each fact, we obtain: (1) Minimal component sets for all answer tokens (2) Importance scores for all attention heads and MLP layers. We set the the greedy threshold to 90\% are are able to recover the original token with 100\% accuracy. After extracting minimal components and importance scores, we compute pairwise similarity measures between all tokens in the short answer and the corresponding-fact tokens in the long answer span, then apply EMD to aggregate these into a single answer-level similarity score.

\rparagraph{Results.}
Figure~\ref{fig:circuit-metrics} reveals that aligned facts exhibit significantly higher mechanistic similarity than misaligned facts across all six metrics (all p < 0.001). This yields direct evidence that behavioral alignment reflects distinct internal mechanisms. 

Set-based metrics show substantial differences: IoU increases by 18.4\% for aligned facts (0.309 vs 0.261), while Containment shows a 12.0\% increase (0.552 vs 0.493). These gaps indicate that factually consistent responses recruit more overlapping components, while inconsistent responses activate divergent computational pathways. Correlation-based metrics reveal complementary patterns. Pearson correlations achieve higher absolute values than Spearman correlations for both attention (0.748 vs 0.686) and MLP components (0.918 vs 0.796), indicating strong linear relationships in raw activation magnitudes. Examining discriminative power, attention metrics show the largest relative gaps regardless of correlation type (Spearman: 14.1\%, Pearson: 14.2\%), suggesting attention patterns are most diagnostic of factual alignment. In contrast, MLP components show divergent patterns: Spearman correlations provide stronger discrimination (8.9\%) than Pearson correlations (3.3\%), indicating that the ranking of MLPs by importance is more informative than their raw activation (i.e., importance)  magnitudes.

Overall, these results confirm our hypothesis: factual alignment corresponds to mechanistic similarity, with aligned facts recruiting more similar computational pathways.

\subsection{Model Internals as Predictors}

Building on our finding that aligned facts exhibit significantly higher mechanistic similarity, we investigate whether these similarity metrics can predict factual alignment. We train a logistic regression classifier using the six similarity metrics as features on a dataset of 360 fact pairs (180 aligned, 180 misaligned) across six models (Llama-3.2 1B/3B, Qwen-2.5 1B/3B, Qwen-3 1.7B/4B). We evaluate performance via 5-fold cross-validation to assess both individual feature contributions and combined predictive power.

\rparagraph{Results and Analysis.}
The predictive modeling results in Figure~\ref{fig:predictive-modeling} validate the mechanistic basis of factual alignment and reveal which similarity metrics best capture this phenomenon. Spearman attention emerges as the strongest individual predictor of alignment (ROC-AUC = 0.86, Accuracy = 0.79), indicating that \textit{ranks} of attention components by importance provide the most reliable measurable signal of factual consistency. The combined feature model achieves robust performance (ROC-AUC = 0.85, Accuracy = 0.80), with logistic regression coefficients revealing the underlying computational structure: Spearman Attention dominates with a coefficient of 1.53, more than four times larger than the next most discriminative feature (Containment).

These results demonstrate that mechanistic interpretability offers not just explanatory insights into model behavior but also a practical tool for predicting factual inconsistencies in LLMs' responses..

\section{Conclusion}

This work establishes factual consistency over query complexity as a critical dimension of LLM reliability. Through \acrnm, we demonstrate that LLMs exhibit systematic misalignment when answering identical factual questions embedded in queries of varying complexity. Our analysis reveals that LLMs are mostly aligned on being factually incorrect and correctness exhibits momentum effects. Our mechanistic analysis provides the first empirical evidence that behavioral factual alignment corresponds to similar internal mechanisms. The predictive modeling results demonstrate practical applications in detecting factual misalignment. Future work should explore factual alignment between short and free-form long-form queries, and investigate targeted interventions on circuit patterns to address the consistency failures identified in LLMs. 

\section*{Limitations}

\rparagraph{\acrnm Dataset and Framework} 
The \acrnm dataset and framework have several limitations. 
(1) The dataset is synthetically generated. While a fully human-annotated dataset would be ideal, it is cost-prohibitive; using LLMs as data generators offers a more balanced quality–resource trade-off. We mitigate this limitation by manually verifying each prompt–answer pair and find that OpenAI’s \texttt{o3-mini-high} is a strong data generator for the \acrnm task (\S\ref{subsec:dataset}). (2) The dataset—and this work—currently covers only English. (3) For evaluation, we adopt the LLM-as-a-judge paradigm to assess factual correctness; in our experiments, however, we find \texttt{Gemini-2.5-Flash} achieves high agreement with human annotations (see \S\ref{sec:benchmarking}). (4) Our dataset and evaluation framework does not address free-form long-form generation.

\rparagraph{Mechanistic Analysis} 
In our mechanistic analysis, we use zero ablation rather than counterfactual activation patching. Although counterfactual activation patching can yield more precise results~\cite{zhangtowards, heimersheim2024use}, constructing counterfactual prompts for complex generation tasks (short- and long-form) is challenging and labor-intensive, especially when relevant facts are distributed across multiple tokens. More importantly, our goal is to demonstrate mechanistic differences between short- and long-form factual retrieval under fact misalignment, which we show in~\S\ref{sec:mechanistic-analysis} using zero ablation.

\section{Acknowledgments}
This work was supported by the Alcatel-Lucent Stiftung and Deutsches Stiftungszentrum through the grant “Equitably Fair and Trustworthy Language Technology” (EQUIFAIR, Grant Nr.
T0067/43110/23). The work of Anne Lauscher is funded under the Excellence Strategy of the German Federal Government and States. The authors gratefully acknowledge the computing time granted by the John von Neumann Institute for Computing (NIC) and provided on the supercomputer JURECA \citep{JURECA} at Jülich Supercomputing Centre (JSC).

\bibliography{main}

\appendix

\section{Appendix}\label{sec:appendix}

\subsection{Dataset Construction} 
\label{subsec:appendix-dataset}

We will release the \acrnm dataset under an open scientific license. 

\rparagraph{Scraping Wikipedia} We collect articles from 15 curated categories. For each category, we use the MediaWiki API to list main-namespace pages and randomly sample candidates. For each candidate, we download the page, extract the main text, and remove tables, images, scripts/styles, hatnotes, navigation boxes, and the table of contents. We collapse whitespace and discard pages shorter than a minimum length of 1000 characters. For every accepted page, we query the Wikimedia Pageviews API to get daily counts over the past 1,095 days and sum them into one popularity score. Within each category, we sort by pageviews and split the list in half to form most and least popular sets.

\rparagraph{Synthetic Data Generation} We generate short/long queries from the scraped Wikipedia text using OpenAI's \emph{o3-mini-high} model (API name \texttt{o3-mini-2025-01-31}, invoked with \texttt{reasoning\_effort=high}). For each article, the prompt instructs the model to: (i) produce \emph{exactly five} self-contained short questions (\texttt{ShortQ1--ShortQ5}) with single-fact, single-answer constraints and \emph{no source referencing}; (ii) give concise answers (\texttt{ShortA1--ShortA5}, each under 10 words) derived \emph{strictly} from the provided text (no parametric knowledge); (iii) compose a long question (\texttt{LongQ}) that \emph{explicitly lists} those five sub-prompts; and (iv) write a fluent long answer (\texttt{LongA}) that synthesizes \emph{only} the five short answers---no extra facts. The instruction emphasizes an ``open-domain phrasing'' test (the subject must be uniquely identifiable outside the Wikipedia article context), bans ambiguous ``pick one of many'' list questions, and enforces a strict, \emph{single-line} output format. The total cost of constructing the dataset via \emph{o3-mini-high} was \textbf{\$18.57}. The prompt for generating the dataset is provided in Table \ref{tab:prompt-dataset-generation} and the dataset schema is provided in Figure \ref{fig:slfa-json-schema}. 

\rparagraph{Topic Categories} Following is a list of all the categories of the topics in the \acrnm dataset: Aesthetics, Algebra, Anthropology, Applied sciences, Artificial intelligence, Astronomy, Cultural studies, Sociology, Software engineering, Spirituality, Statistics, Technology, Telecommunications, Theology, Virtual reality

\subsection{LLM-as-a-judge}
To evaluate LLM-as-a-jduge, we test on 50 samples (250 short questions and 50 long questions) and in total we have 500 atomic facts to be evaluate by LLM. We find LLM to have a an accuracy of 92.\% on short-from and 94\% in long-form responses. 

\rparagraph{Conditions for Incorrect or Correct}:
If any of the following conditions were met, the response was labeled as incorrect: (1) Factual inaccuracy, (2) Semantic dissimilarity, (3) Irrelevance\/Ommission, (4) Contradiction, (5) Hallucinations. For correctness, following conditions have to be met: (1) Direct Semantic Equivalance, (2) Subset/Superset Equivalance, (3) World Knowledge Override (Only used sparsely), (4) Correct Vagueness. These condictions were developed progressively on a validation set of 50 samples and then tested on 50 samples separately. To understand the definition of the each condition, we recommend the reader to go through the prompts in Table \ref{tab:prompts-short-form-eval} and \ref{tab:prompts-long-form-eval}. 

\subsection{Inference}
We generate responses for each short and long query using the Hugging Face API. All models are executed on NVIDIA H100 (80GB) GPU with greedy decoding. The total GPU time required to generate all responses amounts to 192 hours.

We comply with the licensing agreement and adhere to the intended use of each of the open-source and closed-source LLMs. 

\subsection{Mechanistic Interpretability}
We employ the \texttt{NNsight} framework to perform zero ablation on an NVIDIA H100 (80GB) GPU. The total compute time for zero ablation is 768 hours. Subsequently, we manually pair each short-form fact $SQ_k$ with the minimal corresponding span of the fact in the long-form response. This manual pairing process required a total of 16 hours.

\rparagraph{Optimal Transport for Mechanistic Overlap}
We compute mechanistic overlap using optimal transport algorithms, including the Hungarian algorithm and Earth Mover’s Distance (EMD). Both methods yield consistent results—aligned responses exhibit higher mechanistic overlap.

\begin{figure*}[t]
\small
\centering
\begin{minipage}{0.96\textwidth}
\begin{verbatim}
{
  "Category": "History",
  "Topic": "The Punic Wars",
  "URL": "https://en.wikipedia.org/wiki/Punic_Wars",
  "ShortQ1": "When did the First Punic War occur?",
  "ShortA1": "It took place from 264 to 241 BCE.",
  "ShortQ2": "How long did the Punic Wars last?",
  "ShortA2": "They lasted a total of 43 years.",
  "ShortQ3": "Which two powers fought in the Punic Wars?",
  "ShortA3": "Rome and Carthage.",
  "ShortQ4": "Who was the renowned Carthaginian general in the Second Punic War?",
  "ShortA4": "Hannibal Barca was the famous general.",
  "ShortQ5": "What did Hannibal famously cross Italy with?",
  "ShortA5": "Hannibal crossed Italy with war elephants.",
  "LongQ"  : "Discuss the Punic Wars by covering (1) when the First Punic War occurred, 
             (2) how long the Punic Wars lasted, (3) which two powers fought, 
             (4) who the renowned Carthaginian general in the Second Punic War was,
             and (5) what he famously crossed Italy with.",
  "LongA"  : "The First Punic War took place from 264 to 241 BCE,
            and the three Punic Wars altogether lasted about 43 years between
            Rome and Carthage. In the Second Punic War, Hannibal Barca rose as
            the famed Carthaginian commander, becoming legendary for crossing Italy with war elephants.",
  "Pageviews": ...,
  "ShortQ1_Entity": 1,
  "ShortQ2_Entity": 0,
  "ShortQ3_Entity": 1,
  "ShortQ4_Entity": 1,
  "ShortQ5_Entity": 1
}
\end{verbatim}
\end{minipage}
\caption{Example SLFA dataset entry instantiated for the Punic Wars.}
\label{fig:slfa-json-schema}
\end{figure*}

\begin{table*}[t]
\scriptsize
\centering
\begin{tabular}{@{}p{\textwidth}@{}}
\hline
\textbf{Prompt} \\
\hline \hline
{\raggedright\ttfamily
Given a *reference article* (plain text), generate **exactly**:\\
1) **Five** short Q/A pairs: `ShortQ1--ShortQ5`, `ShortA1--ShortA5`  \\
2) **One** long question: `LongQ` (explicitly lists the five sub-prompts)  \\
3) **One** long answer: `LongA` (coherent synthesis of the five short answers)\\
\#\# Examples (compressed)\\
- **Example 1 (placeholder)**  \\
\ \ \ *Reference article:* `[...Punic Wars summary...]`  \\
\ \ \ *Output (abbrev):*  \\
\ \ \ `ShortQ1: When did the First Punic War occur?`  \\
\ \ \ `ShortA1: 264--241 BCE.`  \\
\ \ \ `...`  \\
\ \ \ `LongQ: Discuss by covering (1) \dots (2) \dots (3) \dots (4) \dots (5) \dots`   \\
\ \ \ `LongA: [Single paragraph formed only from ShortA1--5].`\\
- **Example 2 (placeholder)**  \\
\ \ \ *Reference snippet:* `[.....]`  \\
\ \ \ *Output (abbrev):* similar structure to Example 1.\\
\#\# Critical Instructions (strict)\\
\ \ 1.\ \ **Absolute Grounding in Provided Text:** ALL answers (ShortA1-5, LongA) MUST be derived *exclusively* from the information present in the reference text provided below. DO NOT use any external knowledge or information not explicitly stated in the text.\\
\ \ 2.\ \ **Self-Contained \& Precise Questions:** Each short question (ShortQ1-5) must be specific, unambiguous, and fully understandable on its own without needing context from other questions or the article title.\\
\ \ \ \ \ *Precision Example:* Use ``When did the First Punic War occur?'' instead of the vague ``When did the war occur?''.\\
\ \ 3.\ \ **No Source Referencing in Questions:** Questions MUST NEVER refer to the provided text itself. Avoid phrases like ``According to the article...'', ``What does the text mention about...'', ``Which item listed...''. Frame questions as standalone, open-domain factual queries.\\
\ \ 4.\ \ **Strict Single-Fact \& Single-Answer Rule (MOST IMPORTANT):** This constraint is paramount and must be strictly enforced for each ShortQ/ShortA pair:\\
\ \ \ \ \ **One Specific Fact:** Each ShortQ must ask for *one single, specific piece of information*.\\
\ \ \ \ \ **Only One Correct Answer (within the text):** Critically, based *solely* on the provided reference text, there must be *only one possible correct answer* to the ShortQ.\\
\ \ \ \ \ **Mandatory Verification:** Before finalizing any ShortQ, you MUST verify that no other statement or detail *within the provided text* could also serve as a correct answer to that specific question.\\
\ \ \ \ \ **AVOID Ambiguity from Lists/Examples:** If the text presents multiple examples, types, reasons, methods, individuals within a category, etc.\ (e.g., ``Art mediums include painting, digital tools, and ink,'' or ``Key figures were X, Y, and Z''), you MUST NOT formulate a ShortQ asking for *one* of them (e.g., DO NOT ask ``What is *one* art medium mentioned?'' or ``Name *an* important figure.''). Such questions inherently violate the single-answer rule because the text itself provides multiple valid options in that context.\\
\ \ \ \ \ +\ \ **Ensure Subject Uniqueness (Open Domain Test):** The specific entity, event, concept, person, or work being asked about in the question MUST be identifiable *without ambiguity* even when considered outside the context of the source article. Ask yourself: ``If this question were encountered alone, would the subject be clear?''\\
\ \ \ \ \ +\ \ \ \ \ -\ \ **INVALID Example:** ``Which organization funded Short et al.'s work?'' is INVALID if ``Short et al.'s work'' is not a globally famous, uniquely identifiable publication/project (like ``Einstein's theory of relativity''). It improperly relies on the implicit context (``the work mentioned in this article'').\\
\ \ \ \ \ +\ \ \ \ \ -\ \ **VALID Example:** ``What year was the Treaty of Versailles signed?'' is VALID because ``The Treaty of Versailles'' is a globally unique and identifiable historical event.\\
\ \ \ \ \ +\ \ \ \ \ -\ \ **Guideline:** Avoid questions where the subject is a vague reference (e.g., ``the study's findings'', ``their main conclusion'', ``Smith's 2020 paper'' unless that specific paper is uniquely identifiable globally).\\
\ \ \ \ \ **Target Suitable Facts:** Focus ShortQs on unique identifiers (e.g., the *specific name* of the *first* person to do X), distinct dates/years associated with singular events, uniquely defined terms (like *Pax Romana*, if defined as a singular concept in the text), precise numerical values or quantities tied to a specific context, or the outcome of a specific, singular event described, **ensuring the subject meets the Open Domain Test above.**\\
\ \ 5.\ \ **Concise Short Answers:** Each ShortA (ShortA1--ShortA5) must directly state the single fact requested by its corresponding ShortQ, using fewer than 10 words.\\
\ \ 6.\ \ **Structured Longform Composition:**\\
\ \ \ \ \ **LongQ Construction:** The LongQ must explicitly integrate the five preceding ShortQs, typically by listing them as points to be covered (e.g., ``Discuss [Topic] by addressing: (1) [Content of ShortQ1], (2) [Content of ShortQ2]\ldots'').\\
\ \ \ \ \ **LongA Synthesis:** The LongA must synthesize *only* the five ShortAs (ShortA1--ShortA5) into a single, coherent, natural-sounding paragraph. It must flow well and not sound robotic. DO NOT introduce any facts or details not present in the ShortAs. Avoid using bullet points or numbered lists in the LongA.\\
\ \ 7.\ \ **Strict Output Format:** Produce the entire output as a single block of text, strictly adhering to the following CSV-friendly format on one line. Use commas as separators between key-value pairs. Include NO extra text, commentary, introductions, or line breaks before or after this formatted string:\\
\#\#\# Provide One-Line Output Template\\
"ShortQ1 : <value>, ShortA1 : <value>, ShortQ2 : <value>, ShortA2 : <value>, ShortQ3 : <value>, ShortA3 : <value>,\\
ShortQ4 : <value>, ShortA4 : <value>, ShortQ5 : <value>, ShortA5 : <value>,\\
LongQ : <value>,LongA : <value>"\\
**Reference Article:** \\
\{text\} \\
--- \\
**Generate Now:**\\
}\\
\hline
\end{tabular}
\caption{Short-Long form query and answer generation prompt for o3-mini-high}
\label{tab:prompt-dataset-generation}
\end{table*}

\begin{table*}[t]
\scriptsize
\centering
\begin{tabular}{@{}p{\textwidth}@{}}
\hline
\textbf{Prompt} \\
\hline \hline
{\raggedright\ttfamily
**Role and Objective:**\\
You are a meticulous AI Fact Adherence Evaluator. Your primary objective is to perform a rigorous, objective assessment of a ``Generated Short Answer'' by comparing it against a ``Ground Truth Short Answer'' in the context of a specific ``Question''. Your evaluation must determine if the ``Generated Short Answer'' accurately and meaningfully conveys the same information as the ``Ground Truth Short Answer'' or provides an equally valid factual answer to the ``Question'', paying close attention to any specificity implied by the question and ground truth.\\
Core Principles for Evaluation:\\
1. Objectivity: Base your evaluation strictly on the provided information and the criteria outlined below. Avoid personal biases or assumptions.\\
2. Factual Accuracy: The paramount consideration is whether the ``Generated Short Answer'' is factually correct in relation to the ``Question'' and ``Ground Truth''.\\
3. Semantic Meaning: Focus on the core meaning and informational content.\\
4. Contextual Relevance: The ``Generated Short Answer'' must directly and relevantly answer the ``Question''.\\
Input Data Structure:\\
You will be provided with the following for each evaluation task:\\
* Question (ShortQ): SHORT-QUESTION (The specific query the answer should address)\\
* Ground Truth Short Answer (ShortA): SHORT-ANSWER (The pre-validated correct answer, which also sets the expected level of specificity for certain types of questions)\\
* Generated Short Answer (GeneratedShortA): GENERATE-SHORT-ANSWER (The answer to be evaluated)\\
Detailed Evaluation Criteria and Scoring (Output 0 or 1):\\
A. Score 1 (Correct) if ANY of the following conditions are met:\\
1.\ \ Direct Semantic Equivalence:\\
\ \ \ \ * The ``Generated Short Answer'' conveys the same essential information as the ``Ground Truth Short Answer'' and accurately answers the ``Question''.\\
\ \ \ \ * Differences in phrasing, sentence structure, or the use of synonyms are acceptable as long as the core meaning is preserved.\\
2.\ \ Subset/Superset Equivalence (Strict Application Regarding Specificity):\\
\ \ \ \ * If GeneratedShortA is a more specific (subset) version of ShortA (e.g., ShortA: ``Dog'', GeneratedShortA: ``Labrador Retriever dog''), it can be correct if it still fundamentally answers ShortQ accurately and doesn't introduce inaccuracies.\\
\ \ \ \ * If GeneratedShortA is a more general (superset) version of ShortA, it can be correct ONLY IF:\\
\ \ \ \ \ \ \ \ * It still fundamentally and accurately answers ShortQ.\\
\ \ \ \ \ \ \ \ * It doesn't introduce inaccuracies or change the core fact(s) required to answer ShortQ.\\
3.\ \ World Knowledge Override (Strict Application - Use Sparingly):\\
\ \ This applies ONLY IF:\\
\ \ \ \ * The ``Generated Short Answer'' is factually incorrect OR insufficiently specific when compared directly to the ``Ground Truth Short Answer'' based on the criteria above.\\
\ \ \ \ * AND The ``Generated Short Answer'' is a demonstrably true, widely accepted, and commonly known fact that also correctly, directly, and with appropriate specificity answers the ``Question''.\\
\ \ \ \ * AND The ``Generated Short Answer'' is not a niche, controversial, or overly obscure fact.\\
\ \ Checklist before applying World Knowledge Override:\\
\ \ \ \ 1.\ \ Does GeneratedShortA directly and unambiguously answer ShortQ with the necessary specificity? (If no, score 0)\\
\ \ \ \ 2.\ \ Is GeneratedShortA factually true based on broad, verifiable common knowledge? (If no, score 0)\\
\ \ \ \ 3.\ \ If ShortQ demands specificity, is GeneratedShortA a better or equally valid specific answer to that demand than ShortA? (If no, or if ShortA's specificity is contextually more appropriate, score 0)\\
\ \ \ \ 4.\ \ Does GeneratedShortA introduce ambiguity or miss critical nuances that ShortA captures, especially regarding specificity? (If yes, score 0)\\
\ \ Example:\\
\ \ \ \ ShortQ: Who is considered the primary inventor of the telephone?\\
\ \ \ \ ShortA: Alexander Graham Bell.\\
\ \ \ \ GeneratedShortA: Antonio Meucci conceived the telephone first. (Score: 1, IF the LLM's world knowledge strongly supports Meucci as a more accurate answer to ``primary inventor'' despite Bell's common association, and this is a well-established historical correction. This is a high bar.)\\
4.\ \ Correct Vagueness:\\
\ \ * Sometimes, the answer can be correct but vague. For example, if a question says 'Into what must a geometric shape be divided to be symmetric?', the ground truth answer is 'Two or more identical pieces', the generated answer is 'A shape must be divided into two halves'. The generated answer here is correct.\\
\ \ * Similarly, for a technical question, the question could be 'How is the fractal-like shape obtained?', Ground-truth answer 'Finite subdivision rule' and the Generated answer here 'Fractals are created by repeating a pattern at different scales.' is correct here. It is not exactly but the meaning of both is the same.\\
\ \ * For non-technical questions, like 'What is literary criticism?', The ground truth is 'study, evaluation, and interpretation of literature' and the generated answer is 'Literary criticism is the analysis and interpretation of written works.'. The generated answer here is correct. \ \ \ \ \ \ \ \ \ \ \ \ \ \ \ \ \ \ \ \ \ \ \ \\
\ \ * Partial Correction: If the answer is partially correct, Apply World Knowledge Override and see if it can be correct FOR the question. If so, it is correct.\\
B. Score 0 (Incorrect) if ANY of the following conditions are met:\\
1.\ \ Factual Inaccuracy: The ``Generated Short Answer'' is factually incorrect.\\
2.\ \ Semantic Dissimilarity: The ``Generated Short Answer'' conveys a different meaning.\\
3.\ \ Irrelevance: The ``Generated Short Answer'' does not answer the ``Question''.\\
4.\ \ Contradiction: The ``Generated Short Answer'' contradicts the ``Ground Truth Short Answer'' and does not meet the stringent criteria for ``World Knowledge Override''.\\
5.\ \ Hallucination: The ``Generated Short Answer'' introduces general knowledge, which alters the answer's validity and the hallucination is severe.\\
For specificity, you have to judge the Question and see if it requires the answer to be exact and specific. These are often scientific, historical questions where there is only 1 correct answer. If the questions expects a specific answer, only the most closely related generated answer should be correct.\\
\#\#\# OUTPUT FORMAT\\
Return ONE character only: 1 or 0.\\
\#\#\# INPUT\\
Question: \{q\}\\
Ground-truth: \{gt\}\\
Candidate: \{cand\}\\
\textless END\_PROMPT\textgreater\\
}\\
\hline
\end{tabular}
\caption{Prompt for evaluating short-form answers against ground-truth short-form answers.}
\label{tab:prompts-short-form-eval}
\end{table*}

\begin{table*}[t]
\scriptsize
\centering
\begin{tabular}{@{}p{\textwidth}@{}}
\hline
\textbf{Prompt} \\
\hline \hline
{\raggedright\ttfamily
***Role and Objective:***\\
You are an AI Comprehensive Answer Evaluator. Your task is to dissect a ``Generated Long Answer'' and meticulously assess its coverage and accuracy concerning five distinct sub-facts, each defined by a ``Short Question'' and its ``Ground Truth Short Answer''. The ``Generated Long Answer'' is intended to synthesize these five pieces of information. You must pay close attention to any specificity implied by each sub-question and its corresponding ground truth.\\
**Core Principles for Evaluation:**\\
1.\ \ Objectivity: Base your evaluation strictly on the provided information and the criteria outlined below. Avoid personal biases or assumptions.\\
2.\ \ Factual Accuracy: The paramount consideration is whether the ``Generated Short Answer'' is factually correct in relation to the ``Question'' and ``Ground Truth''.\\
3.\ \ Semantic Meaning: Focus on the core meaning and informational content.\\
4.\ \ Contextual Relevance: The ``Generated Short Answer'' must directly and relevantly answer the ``Question''.\\
\#\#\# SUB-QUESTIONS \& GT\\
1.\ \{q1\}\\
\ \ \ \ $\rightarrow$ GT-1: \{a1\}\\
2.\ \{q2\}\\
\ \ \ \ $\rightarrow$ GT-2: \{a2\}\\
3.\ \{q3\}\\
\ \ \ \ $\rightarrow$ GT-3: \{a3\}\\
4.\ \{q4\}\\
\ \ \ \ $\rightarrow$ GT-4: \{a4\}\\
5.\ \{q5\}\\
\ \ \ \ $\rightarrow$ GT-5: \{a5\}\\
\#\#\# CANDIDATE LONG ANSWER\\
\{cand\_long\}\\
**Detailed Evaluation Task and Scoring (List of 5 scores [0 or 1]):**\\
For EACH of the 5 Sub-Facts (iterate from Sub-Fact 1 to Sub-Fact 5):\\
1.\ \ **Isolate Focus:** Concentrate on the current Sub-Fact i (defined by ShortQ[i] and ShortA[i]).\\
2.\ \ **Locate Relevant Information:** Scrutinize the ``Generated Long Answer'' to identify the sentence(s) or phrase(s) that attempt to address ShortQ[i].\\
\ \ \ \ * If no part of ``Generated Long Answer'' appears to address ShortQ[i], assign a score of 0 for this sub-fact and move to the next.\\
3.\ \ **Evaluate Located Information:** If relevant information is found, compare it against ShortA[i] using the following criteria, which mirror the detailed logic of the Short Answer Evaluation:\\
A. Score 1 (Correct) if ANY of the following conditions are met:\\
1.\ \ Direct Semantic Equivalence:\\
\ \ \ \ * The ``Generated Short Answer'' conveys the same essential information as the ``Ground Truth Short Answer'' and accurately answers the ``Question''.\\
\ \ \ \ * Differences in phrasing, sentence structure, or the use of synonyms are acceptable as long as the core meaning is preserved.\\
2.\ \ Subset/Superset Equivalence (Strict Application Regarding Specificity):\\
\ \ \ \ * If GeneratedShortA is a more specific (subset) version of ShortA (e.g., ShortA: ``Dog'', GeneratedShortA: ``Labrador Retriever dog''), it can be correct if it still fundamentally answers ShortQ accurately and doesn't introduce inaccuracies.\\
\ \ \ \ * If GeneratedShortA is a more general (superset) version of ShortA, it can be correct ONLY IF:\\
\ \ \ \ \ \ \ \ * It still fundamentally and accurately answers ShortQ.\\
\ \ \ \ \ \ \ \ * It doesn't introduce inaccuracies or change the core fact(s) required to answer ShortQ.\\
3.\ \ World Knowledge Override (Strict Application - Use Sparingly):\\
\ \ Applies ONLY IF:\\
\ \ \ \ * The ``Generated Short Answer'' is factually incorrect OR insufficiently specific when compared directly to the ``Ground Truth Short Answer''.\\
\ \ \ \ * AND The ``Generated Short Answer'' is a demonstrably true, widely accepted, and commonly known fact that also correctly, directly, and with appropriate specificity answers the ``Question''.\\
\ \ \ \ * AND The ``Generated Short Answer'' is not a niche, controversial, or overly obscure fact.\\
\ \ Checklist:\\
\ \ \ \ 1.\ \ Does GeneratedShortA directly and unambiguously answer ShortQ with the necessary specificity? (If no, score 0)\\
\ \ \ \ 2.\ \ Is GeneratedShortA factually true based on broad, verifiable common knowledge? (If no, score 0)\\
\ \ \ \ 3.\ \ If ShortQ demands specificity, is GeneratedShortA a better or equally valid specific answer than ShortA? (If no, or if ShortA's specificity is more appropriate, score 0)\\
\ \ \ \ 4.\ \ Does GeneratedShortA miss critical nuances that ShortA captures? (If yes, score 0)\\
\ \ Example:\\
\ \ ShortQ: Who is considered the primary inventor of the telephone?\\
\ \ ShortA: Alexander Graham Bell.\\
\ \ GeneratedShortA: Antonio Meucci conceived the telephone first. (Score: 1, IF world knowledge supports Meucci as more accurate.)\\
4.\ \ Correct Vagueness:\\
\ \ * Sometimes the generated answer is correct but vague (e.g., Question: 'Into what must a geometric shape be divided to be symmetric?', ShortA: 'Two or more identical pieces', Generated: 'Two halves' $\rightarrow$ correct).\\
\ \ * Similar logic applies for technical and non-technical contexts, as long as meaning is preserved.\\
\ \ * Partial Correction: If partially correct, apply World Knowledge Override to decide.\\
B. Score 0 (Incorrect) if ANY of the following hold:\\
1.\ \ Factual Inaccuracy.\\
2.\ \ Semantic Dissimilarity.\\
3.\ \ Irrelevance.\\
4.\ \ Contradiction not justified by World Knowledge Override.\\
5.\ \ Severe Hallucination.\\
For specificity, judge whether the Question expects an exact and specific answer (common in science/history). If so, only the most precise matching Generated answer should be correct.\\
**Handling Complexities:**\\
* Information may be split across sentences.\\
* Do not penalize answer order; evaluate each fact independently.\\
* Prefer explicit statements. If heavily implied, err toward 0 unless undeniable.\\
\#\#\# OUTPUT FORMAT\\
Return exactly a JSON list of 5 ints, e.g.\ [1,0,1,1,0]\\
}\\
\hline
\end{tabular}
\caption{Prompt for evaluating long-form answers against five short-form ground truth facts.}
\label{tab:prompts-long-form-eval}
\end{table*}

\begin{table*}[t]
\centering
\begin{tabular}{@{}p{\textwidth}@{}}
\hline
\textbf{Instruction for Short-form QA} \\
\hline \hline
{\raggedright\ttfamily
Answer the question with factual single sentence response for the Topic: {topic}. Question: {question}
}\\
\hline
\textbf{Instruction for Long-form QA} \\
\hline \hline
{\raggedright\ttfamily
Answer to the question should answer everything in the question in a clear and concise manner.Question: {long\_question}
}\\
\hline
\end{tabular}
\caption{Instruction for short and long-form QA.}
\label{tab:instruct-short-long}
\end{table*}

\end{document}